%% file: ms.tex
\documentclass[letterpaper, 10 pt, conference]{ieeeconf}  

\IEEEoverridecommandlockouts                              
\usepackage[pdftex]{graphicx}
\usepackage{epstopdf}
\usepackage[dvipsnames]{xcolor}
\usepackage[caption=false,font=footnotesize]{subfig}
\usepackage{amsmath} 
\usepackage{amssymb}
\usepackage{algpseudocode}
\usepackage{lipsum}
\usepackage{multirow}
\usepackage{tabularx}
\usepackage{algorithmicx}
\usepackage{hyperref}
\usepackage{cite}
\usepackage{setspace}
\usepackage{algorithm}
\usepackage{subfig}

\usepackage{makecell}
\usepackage{array,multirow}

\usepackage{verbatim}

\def\Vec#1{\!\!\hbox{$#1$\kern-0.38em\lower0.85em\hbox{$\vec{}\,$}}\,}%
\newcommand{\bbm}{\begin{bmatrix}}
\newcommand{\ebm}{\end{bmatrix}}

\newcommand{\mb}[1]{\mathbf{#1}}

\newcommand{\state}{\mb{z}}
\newcommand{\fullstate}{\boldsymbol\nu}
\newcommand{\stateSet}{\mathcal{S}}
\newcommand{\ctrl}{\mb{u}}
\newcommand{\ctrlSet}{\mathcal{U}}
\newcommand{\sysState}{\mb{s}}
\newcommand{\actState}{\boldsymbol\xi}

\newcommand{\moutput}{g}
\newcommand{\mOutput}{\mb{g}}
\newcommand{\mfeature}{\mb{x}}

\newcommand{\dweight}{l}  
\newcommand{\dweightMtx}{\mb{L}}  
\newcommand{\mweight}{\mb{w}}

\newcommand{\WeightedDataset}{\mathcal{D}^l}

\title{\textbf{
Learn Fast, Forget Slow: Safe Predictive Learning Control for Systems with Unknown and Changing Dynamics Performing Repetitive Tasks
}}

\author{Christopher D. McKinnon and Angela P. Schoellig
\thanks{The authors are with the Dynamic Systems Lab (www.dynsyslab.org) at the University of Toronto Institute for Aerospace Studies (UTIAS), Canada. email: {\small chris.mckinnon@mail.utoronto.ca}, {\small schoellig@utias.utoronto.ca}} }

\begin{document}

\maketitle
\thispagestyle{empty}
\pagestyle{empty}

\begin{abstract}

We present a control method for improved repetitive path following for a ground vehicle that is geared towards long-term operation where the operating conditions can change over time and are initially unknown. We use weighted Bayesian Linear Regression (wBLR) to model the unknown dynamics, and show how this simple model is more accurate in both its estimate of the mean behaviour and model uncertainty than Gaussian Process Regression and generalizes to novel operating conditions with little or no tuning. In addition, wBLR allows us to use \emph{fast adaptation} and \emph{long-term learning} in one, unified framework, to adapt quickly to new operating conditions and learn repetitive model errors over time. This comes with the added benefit of lower computational cost, longer look-ahead, and easier optimization when the model is used in a stochastic Model Predictive Controller (MPC). In order to fully capitalize on the long prediction horizons that are possible with this new approach, we use Tube MPC to reduce the growth of predicted uncertainty.  We demonstrate the effectiveness of our approach in experiment on a 900\,kg ground robot showing results over 3.0\,km of driving with both physical and artificial changes to the robot's dynamics. All of our experiments are conducted using a stereo camera for localization.

\end{abstract}

\input{src/introduction.tex}

\input{src/related_work.tex}

\input{src/problem_statement.tex}

\input{src/methodology.tex}

\input{src/experiments.tex}

\section{CONCLUSIONS}

In this paper, we have proposed a new method for long-term, safe learning control based on local, weighted BLR. This method is computationally inexpensive which enables fast model updates and allows us to leverage large amounts of data gathered over previous traverses of a path. This enables both fast adaptation to new scenarios and high-accuracy tracking in the presence of repetitive model errors. The model parameters can be determined reliably online which enables our method to be applied in a wide range of operating conditions with little to no tuning. We have demonstrated the effectiveness of the proposed approach in a range of challenging, off-road experiments. We encourage the reader to watch our video at \url{http://tiny.cc/fast-slow-learn} showing the experiments and datasets used in this paper.


\bibliographystyle{unsrt}
\bibliography{bib/bib.bib}

\end{document}

%% file: src/introduction.tex
\section{INTRODUCTION}

This paper presents a new probabilistic method for modelling robot dynamics geared towards stochastic Model Predictive Control (MPC) and repetitive path following tasks. The goal of our approach is to enable a robot to operate in challenging and changing environments with minimal expert input and prior knowledge of the operating conditions.  Our study is motivated by our previous work with Gaussian Processes (GPs) on this topic \cite{McKinnon2018ExpRec}  and an interest in deploying robots in a wide range of operating conditions. Our method requires the unknown part of the dynamics to be linear in a set of model parameters.

Safe control methods have emerged as a way to guarantee that safety constraints (e.g. a bound on maximum path tracking error) are kept in the face of model errors. Having an accurate estimate of model error is of critical importance to the validity of these safety guarantees. In order to derive models for complex systems or systems operating in challenging operating conditions, researchers increasingly rely on tools from machine learning. In particular, probabilistic models are used since they provide a measure of model uncertainty which can naturally be used to derive an upper bound on model error. Two common methods for doing this are GP regression \cite{Zeilinger2017Cautious, McKinnon2018ExpRec, TomlinReachabilityGPs2014} and various forms of local linear regression \cite{jamone2014incremental, Schaal2008BayesEmpLocLin, desaraju2017experience}. 


\begin{figure}[h]
	\centering
	\includegraphics[width=0.5\textwidth]{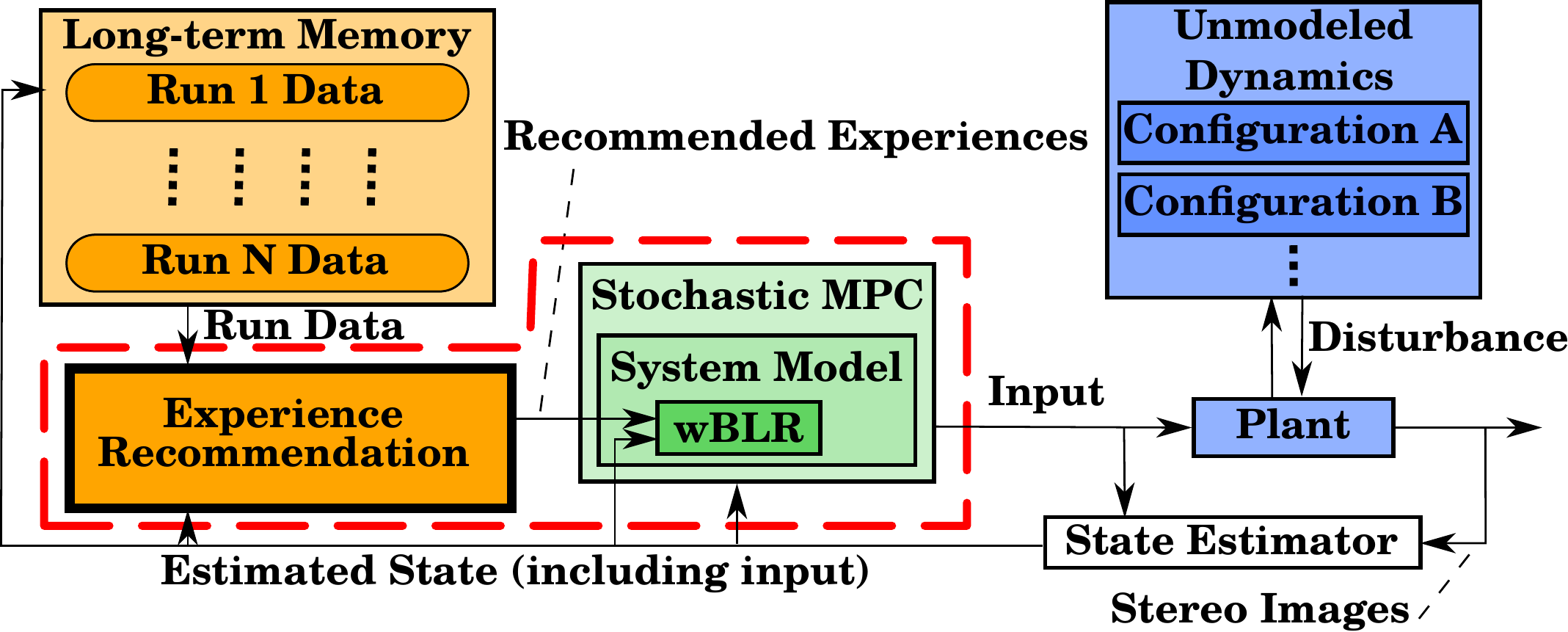}
	\vspace{-5pt}
	\caption{Block diagram showing the proposed model learning method in closed-loop with a safe controller (red dashed box). The system dynamics can change from one run to another and over the course of a run. We use weighted Bayesian Linear Regression (wBLR) to learn the actuator dynamics of the plant. This approach, which enables fast adaptation and long-term learning, is shown to be highly effective in experiment. We encourage the reader to watch our video showing the experiments and datasets used in this paper: \url{http://tiny.cc/fast-slow-learn}.}
	\label{fig:BlockDiagram}
\end{figure}


In our previous work \cite{McKinnon2018ExpRec}, we used GPs to learn the robot dynamics in a number of different operating conditions by leveraging experience gathered over multiple traverses of a path. However, we found that they have a number of limitations that make them difficult to apply in a wide range of operating conditions. First, they are computationally expensive, which limits the number of training points that can be used in the model for control \cite{McKinnon2018ExpRec}. This limits the region of the input space over which the GP is accurate. Second, using maximum likelihood optimization to identify hyperparameters offline did not always result in good closed loop performance. For this reason, we used a fixed set of hyperparameters which limited the range of operating conditions where the learning was effective. Third, given fixed hyperparameters, the GP assumes that the unknown dynamics are globally homoscedastic even though we only fit the model locally along the path. This further limits the effectiveness of a GP-based approach.

In this paper, we propose a new approach to address these limitations: we use weighted Bayesian Linear Regression (wBLR) to model part of the robot dynamics locally along the path (see Fig. \ref{fig:BlockDiagram}). A wBLR model is computationally inexpensive to fit and evaluate. This enables us to use more previous experience to learn repetitive model errors and current experience to adapt quickly to novel operating conditions. We leverage the fact that we are doing a repetitive path following task and a predictive control strategy to efficiently partition past data for fitting our local model. Our approach does not otherwise depend on hyperparameters which, in addition to its relatively simple parametric form, makes it very data efficient and thus able to adapt quickly and reliably to new operating conditions. Finally, in a special case, wBLR can be designed so that it preserves convexity of the optimization problem solved as part of the MPC-based control strategy. As a result, improvements in the model translate well into improvements in control. In this paper, we also show how the model can be combined with Tube MPC to double the look-ahead horizon of our previous approach to three seconds. 

%% file: src/related_work.tex
\section{RELATED WORK}

This work considers the problem of model learning for repetitive path following and a stochastic MPC. Recent work on this subject can be broadly grouped into three categories depending on how they group data to construct a model for robot dynamics.

First, single mode learning control. This class of methods learns a single model for the robot dynamics. This means that all data gathered by the robot can be grouped into one model and used to train any model parameters and validate them to avoid overfitting. This class of methods has shown impressive results control of ground robots \cite{gao2014tube, Zeilinger2017Cautious}, quadrotors \cite{desaraju2017experience}, manipulators \cite{Schaal2008BayesEmpLocLin} and humanoid robots \cite{SigaudXCSFvsLWPR2011}. This style of approach can learn new dynamics quickly, but if the robot dynamics can change due to a factor that is not included in the model (e.g. snow or wet ground changing the dynamics of ground robot) this class of methods only has the capacity to learn the robot dynamics in one such operating condition. It must either `forget' all previous experience and adapt to the new operating condition from scratch or risk unsafe and sub-optimal behaviour due to model inaccuracy.

To address this, multi-modal learning methods learn a set of models to account for the dynamics in all operating conditions. The number of models in this set may be fixed or grow as new conditions are encountered during robot operation. This class of methods can still leverage all data accumulated in each operating condition to fit and validate complex models. This class of methods has shown impressive results in motion planning to avoid dynamic obstacles \cite{Roy2013DynamicSafePlanning}, repetitive path following \cite{McKinnon2018ExpRec}, and legged robot locomotion \cite{Mouret2017PriorSelection} among others \cite{calandra2015learning, Luders2013robustParafoil, jo2012IMMEKF}. The main drawback of these methods is that they either assume the number of operating conditions is fixed, which presents similar limitations to the single mode methods, or, in the case of \cite{McKinnon2018ExpRec} which was performing repetitive path following, take one full traverse of the path to adapt to new operating conditions rather than adapting to new operating conditions over the course of a run. Adapting quickly to new operating conditions as they arise remains a challenge.

To bridge this gap, recent methods such as \cite{Pereida2018, Abbeel2016OneShot} include both a complex model trained on lots of data with a simple online adaptation term to that can be updated quickly to adapt to new, previously unseen tasks. The simplicity of this online learning term enables fast adaptation to new conditions without worrying about overfitting or gathering sufficient data to do a complex model identification and validation. The long-term learning components, however, remain fixed and it is not clear how to update the long-term learning models efficiently. For example, \cite{Abbeel2016OneShot} used a neural network trained on several hours of data and then fixed as the long-term learning component and linear regression updated recursively based on recent measurements to construct a `fast adaptation' term that also captured the uncertainty in the robot dynamics. In this work, we propose a solution that couples a relatively simple model structure that can be adapted quickly to novel operating conditions with the ability to leverage lots of data gathered over many traverses of the path in various operating conditions. We use local models to achieve high performance with this relatively simple model form, and data weighting to incorporate the most relevant past data to improve from repeated traverses in similar operating conditions. This combines the long-term and fast adaptation components in one, unified, probabilistic framework.

In light of the current approaches and their limitations, the contributions of the paper are (i) to present a model learning framework that supports \textit{fast adaptation}, \textit{long-term learning}, and is tailored to predictive control; (ii) to incorporate that model (and its model uncertainty estimate) in a stochastic predictive control scheme; and (iii) to demonstrate the advantage of \textit{fast adaptation} and \textit{long-term learning} in path tracking experiments over challenging terrain.

%% file: src/problem_statement.tex
\section{PROBLEM STATEMENT}

The goal of this work is to learn a probabilistic model for the dynamics of a ground robot performing a repetitive task, and show how it can be integrated with a state-of-the-art path following controller for high performance control while maintaining a quantitative measure of safety. The robot may be subjected to changes in its dynamics due to factors such as payload, terrain, or tyre pressure. We assume that these factors cannot be measured directly and all possible disturbances are not known ahead of time. A good algorithm should scale to long-term operation, take advantage of repeated runs in the same operating conditions, and adapt quickly to new operating conditions. The model must include a reasonable estimate of model uncertainty that acts as an upper bound on model error at all times.

We consider systems with dynamics of the form:
\begin{align}
\label{eq:GeneralSystemDynamics}
\mb{s}_{k+1} &= \mb{s}_k + dt \overbrace{\mb{f}(\mb{s}_k, \boldsymbol\xi_{k})}^{known}, \\
\label{eq:GeneralActuatorDynamics}
 \boldsymbol\xi_{k+1} &= \underbrace{\mb{g}^0(\boldsymbol\xi_k, \mb{u}_k)}_{known} + dt\underbrace{\mb{g}_k(\mfeature_k)}_{unknown},
\end{align}
where the state of the system $\mb{s}$ evolves according to known dynamics $\mb{f}(\cdot)$ that depend on $\mb{s}$ and the state of the actuators $\boldsymbol\xi$. We assume that our control input $\mb{u}$ affects the actuator dynamics which consist of a known part $\mb{g}^0(\cdot)$ and an unknown and potentially changing part $\mb{g}_k(\cdot)$ that we wish to learn. The unknown dynamics depend on a feature vector $\mfeature$ that may be, for example, composed of $\actState$ and $\ctrl$ or nonlinear functions of these depending on prior knowledge about the system. The subscript refers to the timestep and $dt$ is the duration of a timestep.

The system is constrained by state and input constraints. Let $\state_k = [\mb{s}_k^T, \actState_k^T]^T$. Then:
\begin{align}
 \state_k \in \stateSet,
 \ctrl_k \in \ctrlSet.
\end{align}

We assume a Gaussian belief over the state at each time step and enforce constraints probabilistically using a chance-constrained formulation so that the probability of violating state and input constraints is kept below an acceptable threshold.  Since enforcing these constraints jointly can lead to undesirable, conservative behaviour, we enforce them individually, see \cite{Zeilinger2017Cautious} for a detailed explanation.

%% file: src/methodology.tex
\section{METHODOLOGY}
In this section, we present our approach for long-term, safe learning control with fast adaptation. Our approach makes extensive use of wBLR to model the system dynamics. We assume a known nonlinear model for the plant with unknown actuator dynamics that are linear in a set of of model parameters. We use wBLR to determine the model parameters and a measure of run similarity to determine the data weights. This allows us to compute the posterior for the model parameters in closed form, avoiding iterative approaches such as \cite{Schaal2008BayesEmpLocLin}, which also optimizes the data weights. We then formulate the control problem as a Tube MPC problem following work in \cite{Zeilinger2017Cautious, Lam2010ContouringControl} but using a modified ancillary controller.

\input{src/blr.tex}

%
%

\input{src/mpc.tex}

%% file: src/blr.tex
\subsection{Weighted Bayesian Linear Regression}


In this section, we give a brief overview of wBLR, which is used to learn the actuator dynamics, $\mb{g}_k(\cdot)$. It is an extension of Bayesian linear regression (BLR), as presented in \cite{MurphyProbPerspective}, and a modification of \cite{Schaal2008BayesEmpLocLin}, where we assume a data weighting is obtained in a separate step.  

We consider each dimension of $\mb{g}_k(\cdot)$ separately. For this section, we will refer to a single dimension of $\mb{g}_k(\cdot)$ as $g(\cdot)$. For a given $\mfeature_k$ the corresponding sample for $g(\mfeature_k)$, denoted as $g_k$, may be calculated as $g_k=(\xi_{k+1} - g^0(\actState_k,\ctrl_k))/dt$, where $\xi_{k+1}$ and $g^0(\cdot)$ are the relevant dimensions of $\actState_{k+1}$ and $\mb{g}^0(\cdot)$, respectively.

Suppose we are given a \emph{weighted} dataset $\WeightedDataset~=~\left\lbrace\mb{x}_i,~\moutput_i,~\dweight_i \right\rbrace_{i=1}^n$ with scalar weights $\dweight_i \in [0,1]$ that determine the importance of each data point. If $\dweight_i=0$, the point has no influence on the regression, and if $\dweight_i=1$, the point is fully included. In a simple scenario, all weights can be set to $1$, in which case we recover regular BLR. We assume that the dynamics of interest depend on a vector of model parameters $\mb{w}$ and are of the form
\begin{align}
\label{eq:GeneralBLRModel}
  g(\mfeature) = \mb{w}^T\mb{x} + \eta,
\end{align}
where $ \eta ~\sim \mathcal{N}(0, \sigma^2)$. The goal of wBLR is to determine the distribution for $\mweight$ and $\sigma^2$ given $\WeightedDataset$.

We start by assuming that each data point is independent and weight the contribution of each point as follows:
\begin{equation}
\label{eq:DataLikelihood}
	p(\mOutput\,|\,\mb{X},\mb{w},\sigma^2) = \prod_{i=1}^n \mathcal{N}(g_i\,|\,\mweight^T\mfeature_i, \sigma^2)^{\dweight_i},
\end{equation}
where $\mOutput$ is a vector of stacked $g_i$, and $\mb{X}$ is a matrix with rows $\mb{x}_i^T$. The intuition is one point raised to $l_i=2$ would have the same contribution as two identical points and two identical points with $l_i=0.5$ would have the same contribution as one data point. To avoid over-confident estimates, we restrict $l_i~\in~[0,1]$. With this likelihood, the conjugate prior is a Normal Inverse Gamma ($NIG$) distribution \cite{MurphyProbPerspective} which gives us the following priors for $\mb{w}$ and $\sigma^2$:
\begin{align}
  \label{eq:weightsPrior}
  p(\mb{w}|\sigma^2) &\sim \mathcal{N}(\mweight\,|\,\mb{w}_0, \sigma^2\mb{V}_0),\\
  \label{eq:outputVariancePrior}
  p(\sigma^2) &\sim IG(\sigma^2\,|\,a_0, b_0),
\end{align}
where $\mb{w}_0$ is the prior mean for the weights, $\mb{V}_0$ is a prior inverse sum of squares of $\mb{x}$, and $a_0$ and $b_0$ are the parameters of the Inverse Gamma distribution, which are proportional to the effective number of data points in the prior and $a_0$ times the prior output variance.

The likelihood, \eqref{eq:DataLikelihood}, can be manipulated into a $NIG$ distribution over $\mb{w}, \sigma^2$ so that \eqref{eq:weightsPrior} and \eqref{eq:outputVariancePrior} form a conjugate prior and the posterior joint distribution over $\mb{w}$ and $\sigma^2$ is:
\begin{align}
	p(\mb{w}, \sigma^2|\WeightedDataset) &= NIG(\mb{w},\sigma^2|\mb{w}_N,\mb{V}_N,a_N,b_N)\\
		&\triangleq \mathcal{N}(\mb{w}\,|\,\mb{w}_N, \sigma^2\mb{V}_N)IG(\sigma^2\,|\,a_N, b_N),
\end{align}
where,
\begin{align}
  \label{eq:posterior_update_w}
    \mb{w}_N &= \mb{V}_N(\mb{V}_0^{-1}\mb{w}_0 + \mb{X}^T\dweightMtx\mOutput),\\
   \label{eq:poseterior_update_V} 
    \mb{V}_N &= (\mb{V}_0^{-1}+\mb{X}^T\dweightMtx\mb{X})^{-1},\\
    \label{eq:posterior_update_a}
    a_N &= a_0 + tr(\dweightMtx)/2,\\
    \label{eq:posterior_update_b}
    b_N &= b_0 + \frac{1}{2}(\mb{w}_0^T\mb{V}_0^{-1}\mb{w}_0 + \mOutput^T\dweightMtx\mOutput - \mb{w}_N^T\mb{V}_N^{-1}\mb{w}_N),
\end{align}
where $tr(\cdot)$ is the trace operator and $\dweightMtx$ is a diagonal matrix of the data weights $l_i$. The posterior marginals are then:
\begin{align}
	\label{eq:PosteriorMarginalVariance}
	p(\sigma^2|\WeightedDataset) &= IG(\sigma^2\,|\,a_N,b_N),\\
	\label{eq:PosteriorMarginalWeights}
	p(\mb{w}|\WeightedDataset) &= \mathcal{T}(\mweight\,|\,\mb{w}_N,\frac{b_N}{a_N}\mb{V}_N,2a_N)
\end{align}
where $\mathcal{T}$ is a Student t distribution. This gives us all of the components we need to make predictions of the state at future timesteps. It is important to note that while the uncertainty in $\sigma^2$ decreases as more data is added, the mean value for $\sigma^2$ can increase or decrease to reflect the data. The model uncertainty is then passed to the controller. This is in contrast to a GP (with fixed hyperparameters) where the uncertainty only decreases to a value determined by the hyperparameters as data is added. While it is possible to update the hyperparameters for a GP online, this is a computationally expensive operation that scales poorly with the size of the dataset and validating hyperparameters on a sufficiently large dataset is important to avoid overfitting.

\subsubsection{Recursive Updates}
\label{sec:BLRRecursiveUpdate}
When dealing with streaming data such as the data generated by a robot driving, it can be useful to continually update the model with recent data in order to adapt quickly to new scenarios. To do this  while ensuring the model stays flexible enough to adapt to sudden changes, we recursively update the prior parameters while keeping the strength of the prior fixed at a pre-determined value $n_0$. The value of $n_0$ determines how many effective data points we attribute to the prior. A large value for $n_0$ results in smoother estimates for the $\mb{w}$ and $\sigma^2$ while a smaller value for $n_0$ allows them to vary more quickly. If we start with fewer than $n_0$ points in the prior, e.g. $a_0 < n_0/2$, we update the prior using \eqref{eq:posterior_update_w}-\eqref{eq:posterior_update_b} with the weight for the new point set to one, and set the posterior parameters to the prior for the next timestep. Once $a_0$ reaches $n_0/2$, we use \eqref{eq:posterior_update_w}-\eqref{eq:posterior_update_b} with the weight for the new point set to one and then use the following re-weighting to keep $n_0$ constant:
\begin{align}
  \mb{V}_{0^*} &= \frac{n_0 + 1}{n_0}\mb{V}_{N}, \qquad & \mb{w}_{0^*} &= \mb{w}_N,\\
  a_{0^*} &= \frac{n_0}{n_0 + 1} a_{N},  \qquad & b_{0^*} &= \frac{n_0}{n_0 + 1} b_N.
\end{align}
The parameters  $(\cdot)_{0^*}$ are the re-weighted parameters which become the new prior. This is equivalent to assigning the prior \textit{and} the new point a weight of $n_0/(n_0 + 1)$ and carrying out a weighted update using \eqref{eq:posterior_update_w}-\eqref{eq:posterior_update_b}.  
Compared to GPs, this gives us more control on how fast the model adapts. For a GP, a new point must either displace an existing one if the model has fixed size or increase the model size, which increases the computational cost of the model and will make it less flexible over time as more points are added. For wBLR, the influence of old data decreases after each re-weighting. The rate at which this happens depends on $n_0$, which is a parameter of our choosing and does not affect the computational cost of the model.

\subsubsection{Preserving Convexity for MPC}

MPC usually uses a gradient-based solver to compute the optimal control sequence efficiently. It is therefore desirable to maintain properties such as convexity in the optimization problem. Suppose that the MPC optimization problem is convex to begin with (e.g. the objective and inequality constraints are convex and $\mb{f}(\cdot)$ and $\mb{g}^0(\cdot)$ are affine). Then, if $\mb{g}_k(\cdot) $ is affine in $\mb{x}_k$, the new optimization problem will be convex for any choice of $\mweight$. See \cite[Sec.~4.2]{BoydCvxOpt}.


\subsection{Data Management}


\begin{figure}
\centering
\scriptsize
\def\svgwidth{0.475\textwidth}
\graphicspath{{figs/RobochunkPathConcept/}}
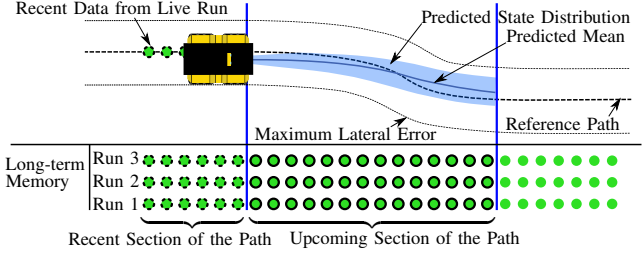
\vspace{-0.25cm}
  \caption{The predicted trajectory (shaded blue) is shown superimposed over the reference path in parallel with the storage structure for data from previous runs (green circles) that is indexed by run and location along the path. Data along the recent section of the path (circles with dotted outlines) is used to estimate the similarity between the current run and each previous run. This similarity is used to weight data from the upcoming section of the path (circles with solid outlines) and construct the predictive model used in MPC. We also use recent data from the current run to recursively update the model and adapt quickly to novel operating conditions and non-repetitive changes. The size of the regions of the path considered \textit{upcoming} and \textit{recent} may be considered hyperparameters that are linked to the MPC problem.}
  \label{fig:ExpRecConceptualDrawing}
\end{figure}

The purpose of our method is to construct the best possible model of the system dynamics for MPC. MPC uses the dynamics over the upcoming section of the path to compute the control input. Referring to Fig. \ref{fig:ExpRecConceptualDrawing}, we use data from the \textit{recent section path} to determine the weights that indicate which runs are most similar to the current run. Given these weights, we use data over the \textit{upcoming} section of path (determined by the MPC look-ahead horizon) to construct a predictive model for the robot dynamics using wBLR. We use two mechanisms to adapt quickly to new scenarios and take advantage of repeated traverses in similar conditions.

\subsubsection{\textbf{Fast Adaptation}}
\label{sec:ShortTermLearning}
In order to adapt quickly to new scenarios, we use the most recent data pair $\{g_i, \mfeature_i\}$ generated by the robot to update the model at every timestep. We use the \emph{recursive update} explained in the previous section. These parameters are used as the prior at each timestep. In our previous work \cite{McKinnon2018ExpRec}, the model reverted to a conservative form when the current dynamics did not match the dynamics in any previous run. While this preserved safety, it took one traverse of the path before the robot could adapt to new conditions. The approach presented in this paper enables the robot to adapt to new conditions as they arise, which is demonstrated in Sec. \ref{sec:ClosedLoopComparison}.

\subsubsection{\textbf{Long-term Learning}}
\label{sec:LongTermLearning}
To improve controller performance in the face of repetitive changes, we leverage data from previous runs in similar operating conditions. We consider data from all previous runs because the model update is efficient and the cost to evaluate the model does not depend on the number of points used to construct it. Let $\mathcal{D}_j^-$ be data from previous run $j$ over the \emph{recent section of the path} (see Fig. \ref{fig:ExpRecConceptualDrawing}) and $\hat{m}_j^-$ be a model constructed from $\mathcal{D}_j^-$. Let $(\cdot)_{i,j}$ refer to point $i$ in run $j$ and let $n$ be the current run. 

\paragraph{Outlier Rejection} First, we check whether using data from each previous run is likely to result in model errors that violate the assumptions of the safe controller. Namely that a given percentile of model uncertainty is a reasonable upper bound for model error. For each previous run, we use $\hat{m}_j^-$ to generate predictions for the mean and variance corresponding to each $\mfeature_{i,n}$ in recent data from the current run. We then compute the Z-score for each prediction given the associated measurement $g_{i,n}$ and compare this to the Z-score associated with the percentile of model error used as an upper bound in MPC (e.g. a Z-score of 2 for the 95th percentile). If the proportion of points outside of this threshold is higher than would be expected by chance (using the binomial test), we reject the run from further consideration. See \cite{McKinnon2018ExpRec} for details.

\paragraph{Weighted Model Update} Now that we have identified runs that will produce a model with valid confidence intervals (for safety), we weight data from each run according to its similarity to the current run (for performance). We compute the posterior probability of model $\hat{m}^-_j$ using:
\begin{equation}
\label{eq:ModelPosterior}
 p(\hat{m}^-_j|\mathcal{D}_n^-) \propto p(\mathcal{D}_n^- | \hat{m}^-_j) p(\hat{m}^-_j).
\end{equation}
The first term on the right is the likelihood of recent data given model $\hat{m}^-_j$.
The second term on the right is the prior, which we assume to be equal for all runs; however, it could be informed by other sources such as computer vision, a weather report, or user input. Similar to our previous work \cite{McKinnon2018ExpRec}, we reject any run that has lower probability than the prior of generating $\mathcal{D}_n^-$. This is to ensure that experience added is likely to improve the performance beyond what could be achieved with no additional experience. 

To update the parameters of the predictive model, we collect data from each previous run over the \emph{upcoming section of the path} and weight each point in run $j$ by $l_{i,j}=p(\mathcal{D}_n^- | \hat{m}^-_j)/ p(\mathcal{D}_n^- | \hat{m}^-_{j^*})$, $i=1..n^+_j$ where $n^+_j$ is the number of points in run $j$ over the \emph{upcoming section of the path} and $j^*$ is the run with maximum posterior probability. This satisfies $\dweight_{i,j} \in\left[ 0, 1\right]$ and means that the effective number of points can increase with each additional run. 

With these weights, we use \eqref{eq:posterior_update_w}-\eqref{eq:posterior_update_b} to compute the posterior parameters of the predictive model. This update (based on data from previous runs) is considered to be location specific and therefore discarded after computing the control; that is, the recursively updated prior becomes the prior for the next timestep.

%% file: figs/RobochunkPathConcept/robochunk_data_with_live_run_MPC.pdf_tex
\begingroup%
  \makeatletter%
  \providecommand\color[2][]{%
    \errmessage{(Inkscape) Color is used for the text in Inkscape, but the package 'color.sty' is not loaded}%
    \renewcommand\color[2][]{}%
  }%
  \providecommand\transparent[1]{%
    \errmessage{(Inkscape) Transparency is used (non-zero) for the text in Inkscape, but the package 'transparent.sty' is not loaded}%
    \renewcommand\transparent[1]{}%
  }%
  \providecommand\rotatebox[2]{#2}%
  \ifx\svgwidth\undefined%
    \setlength{\unitlength}{679.95361328bp}%
    \ifx\svgscale\undefined%
      \relax%
    \else%
      \setlength{\unitlength}{\unitlength * \real{\svgscale}}%
    \fi%
  \else%
    \setlength{\unitlength}{\svgwidth}%
  \fi%
  \global\let\svgwidth\undefined%
  \global\let\svgscale\undefined%
  \makeatother%
  \begin{picture}(1,0.3739854)%
    \put(0,0){\includegraphics[width=\unitlength]{robochunk_data_with_live_run_MPC.pdf}}%
    \put(0.39157216,0.15731621){\color[rgb]{0,0,0}\makebox(0,0)[lb]{\smash{Maximum Lateral Error}}}%
    \put(0.78619926,0.17359553){\color[rgb]{0,0,0}\makebox(0,0)[lb]{\smash{Reference Path}}}%
    \put(0.78761279,0.31301859){\color[rgb]{0,0,0}\makebox(0,0)[lb]{\smash{Predicted Mean}}}%
    \put(0.65344794,0.34261532){\color[rgb]{0,0,0}\makebox(0,0)[lb]{\smash{Predicted State Distribution}}}%
    \put(-0.0009088,0.11025605){\color[rgb]{0,0,0}\makebox(0,0)[lb]{\smash{Long-term}}}%
    \put(0.13640324,0.04789886){\color[rgb]{0,0,0}\makebox(0,0)[lb]{\smash{Run 1}}}%
    \put(0.13640324,0.08437193){\color[rgb]{0,0,0}\makebox(0,0)[lb]{\smash{Run 2}}}%
    \put(0.13640324,0.12084501){\color[rgb]{0,0,0}\makebox(0,0)[lb]{\smash{Run 3}}}%
    \put(0.01521851,0.35615517){\color[rgb]{0,0,0}\makebox(0,0)[lb]{\smash{Recent Data from Live Run}}}%
    \put(0.44620953,-0.0057529){\color[rgb]{0,0,0}\makebox(0,0)[lb]{\smash{Upcoming Section of the Path}}}%
    \put(0.09793799,-0.00746603){\color[rgb]{0,0,0}\makebox(0,0)[lb]{\smash{Recent Section of the Path}}}%
    \put(0.21171097,0.04207038){\color[rgb]{0,0,0}\makebox(0,0)[lb]{\smash{$\underbrace{\hspace{39pt}}$}}}%
    \put(0.38265539,0.04207038){\color[rgb]{0,0,0}\makebox(0,0)[lb]{\smash{$\underbrace{\hspace{93pt}}$}}}%
    \put(-0.00082026,0.08348021){\color[rgb]{0,0,0}\makebox(0,0)[lb]{\smash{Memory}}}%
  \end{picture}%
\endgroup%

%% file: src/mpc.tex
\subsection{Path Following MPC Controller Design}
This section outlines our MPC formulation including the path parametrization, cost function, ancillary control design, and uncertainty propagation. We use a Model Predictive Contouring approach, based on \cite{Lam2010ContouringControl}, which expresses position error as lag error (parallel to the path) and contouring error (perpendicular to the path) and uses a virtual input to drive reference states along the path.

\subsubsection{\textbf{Uncertainty Propagation}}
\label{sec:UncertaintyPropagation}
We assume a Gaussian belief over the state at each time step and nonlinear dynamics for the plant. This allows us to use the Extended  Kalman Filter (EKF) prediction equations to propagate our belief of the state into the future given a series of inputs  \cite{Zeilinger2017Cautious}. We include uncertainty in the full state $\state = [\mb{s}^T, \boldsymbol\xi^T]^T$, the actuator model  parameters $\mweight$, and the actuator model offset $\boldsymbol\eta$. Let $\mb{h}(\cdot)$ be the combined dynamics model \eqref{eq:GeneralSystemDynamics} and \eqref{eq:GeneralActuatorDynamics} and $\mb{A}$ be the Jacobian of $\mb{h}(\cdot)$ with respect to the stacked full state and parameters, $\mb{A}=[\mb{A}_{\state}, \mb{A}_{\mb{w}}]$. The mean $\bar{\state}_k$ and covariance $\mb{\Sigma}^{\mb{\state\state}}_k$ can be updated using:
\begin{align}
\label{eq:EKF_MeanPrediction}
\bar{\state}_{k+1} &= {\mb{h}}(\bar{\state}_k, \ctrl_k),\\
\label{eq:EKF_CovPrediction}
\boldsymbol{\Sigma}^{\state\state}_{k+1} &= \mb{A}\mb{P}_k\mb{A}^T + \mb{Q}_k,\\
\label{eq:EKFPkDefinition}
\mb{P}_k &= \bbm \boldsymbol\Sigma^{\state\state}_{k}& \mb{0}\\ \mb{0}& \boldsymbol\Sigma^{\mb{w}\mb{w}}_{k} \ebm,
\end{align}
where  $\boldsymbol\Sigma^{\mb{w}\mb{w}}_k$ is a block-diagonal matrix containing the model weight covariance matrix from \eqref{eq:PosteriorMarginalWeights} for each dimension of $\mb{g}_k(\cdot)$, $\mb{Q}_k$ is the process noise covariance, and $\ctrl_k$ comes from MPC. The only non-zero components in $\mb{Q}_k$ are the diagonal elements corresponding to uncertainty in the output of the actuators for which we use the posterior mean of the variance from \eqref{eq:PosteriorMarginalVariance}. In this framework, we can include uncertainty in the evolution of the model parameters $\mweight$ by modelling their dynamics as a random walk. In this work, we consider them to be fixed at the posterior estimate over the lookahead horizon.

The predicted uncertainty can be used to compute a confidence  set around the mean prediction that the true system is guaranteed to lie within with high probability. 


\subsubsection{\textbf{Ancillary State Feedback Controller}}


The method for uncertainty propagation in Sec. \ref{sec:UncertaintyPropagation}  but does not take into account the fact that the controller can take corrective actions to reduce the predicted uncertainty \cite{Zeilinger2017Cautious}. The result is that the predicted uncertainty can grow quickly and without bound resulting in conservative control actions \cite{Zeilinger2017Cautious}. A common approach to account for feedback when predicting uncertainty is to use Tube MPC \cite{Tomlin2013RobustLearningMPC} and use an ancillary controller in the predictive model that drives the state towards the predicted mean \cite{Zeilinger2017Cautious}. 

In contrast to other approaches for tube MPC for non-linear systems, we make use of the fact that our actuator dynamics are linear to design linear ancillary controllers for these states. This keeps the uncertainty in these states bounded, which limits the uncertainty \emph{growth} in other states over the prediction horizon. Section \ref{sec:AncillaryCtrlForUnicycle} shows how we apply this to a unicycle-type robot.

\subsubsection{\textbf{Constraint Tightening}}
\label{sec:ConstraintTightening}
Since our predictive model has uncertainty, we must tighten the constraints on the state and input to make sure the true system respects the true constraints (with high probability), and that the ancillary control policy remains feasible for our choice of the inputs. Our treatment of the constraint tightening follows \cite{Zeilinger2017Cautious}.  For contouring error $e^c$, our chance constraints are:
\begin{align}
  p(e^c_k \leq e^{c, max}) &\geq 1 - \epsilon_c\\
  \Leftrightarrow
  \label{eq:TightenedStateConstraints}
  e^c_k + r^c \sqrt{(\mb{t}_k^{\perp})^{T}\boldsymbol\Sigma^{\state\state}_{k}\mb{t}^{\perp}_k} &\leq e^{c,max},
\end{align}
where $r^c$ is the quantile of the Gaussian CDF corresponding to the small probability of violating the contouring constraint $\epsilon_c$ (e.g. 2.0 for $\epsilon_c=0.05$) \cite{Zeilinger2017Cautious}, and $\mb{t}^{\perp}_k$ is a unit vector perpendicular to the path at time $k$. Other constraints on the state may be treated analogously.

Analogous treatment of the input constraints yields:
\begin{align}
    p(u^{[i]}_k < u^{[i], max}) &\leq 1 - \epsilon_{u^{[i]}}\\
  \Leftrightarrow
  u^{[i]}_k + r^{u^{[i]}} K_{u^{[i]}}\sqrt{(\sigma_k^{e})^2} &\leq u^{[i], max},
\end{align}
where $u^{[i]}$ is the $i$th of $\mb{u}$, $K_{u^{[i]}}$ is an associated ancillary gain which acts on an error of our choosing, $e$, and $\sigma^e$ is the standard deviation associated with that error. Here, we can see that while the ancillary controller reduces the prediction uncertainty it will also reduce the control input available for controlling the nominal state.

The feedback gain can be chosen as an infinite horizon LQR controller with the same cost function as MPC \cite{Zeilinger2017Cautious, gao2014tube} or included in the optimization problem \cite{Tomlin2011feedback}, but we found that a wide range of gains worked for our system so left the gain as a tuning parameter.

\subsection{Optimal Control Problem}

At each timestep, we wish to solve for the optimal states and inputs subject to a set of safety constraints derived from the model uncertainty, path tracking error and actuator constraints.  The decision variable is $\fullstate_H = [\mb{u}_0, \mb{z}_1, ... \mb{u}_{N-1}, \mb{z}_N]^T$. This leads to the following optimization problem:
\begin{align}
\label{eq:CostFcnCost}
& \underset{\bar{\fullstate}_H}{\text{minimize}}& & J(\bar{\fullstate}_H) \\
& \text{subject to}
\label{eq:CostFcnDynamicConstraints}
& &  \bar{\state}_{k+i+1} = {\mb{h}}(\bar{\state}_{k+i}, \ctrl_{k+i}, \mfeature_{k+i}), \; i=0..N-1,\\
\label{eq:CostFcnStateConstraints}
& & & p(\state_{k+i+1} \in \stateSet) \geq 1-\boldsymbol\epsilon^{\state}, \; i=0..N-1, \\
\label{eq:CostFcnInputConstraints}
& & & p(\ctrl_{i} \in \ctrlSet) \geq 1-\boldsymbol\epsilon^{\ctrl}, \; i=0..N-1, 
\end{align}
where $\boldsymbol\epsilon^{(\cdot)}$ is a vector of small, acceptable probabilities of violating each state and input constraint, which must be solved at every timestep and $J(\cdot)$ is a quadratic cost that penalizes position, heading, and velocity error, and includes a smoothing term to avoid high frequency inputs. We use the mean of each random variable $(\bar{\cdot})$  to approximate the expected cost and enforce the dynamics constraints.

%% file: src/experiments.tex
\section{APPLICATION TO A GROUND ROBOT}

\begin{figure}
	\centering
	\includegraphics[width=0.5\textwidth]{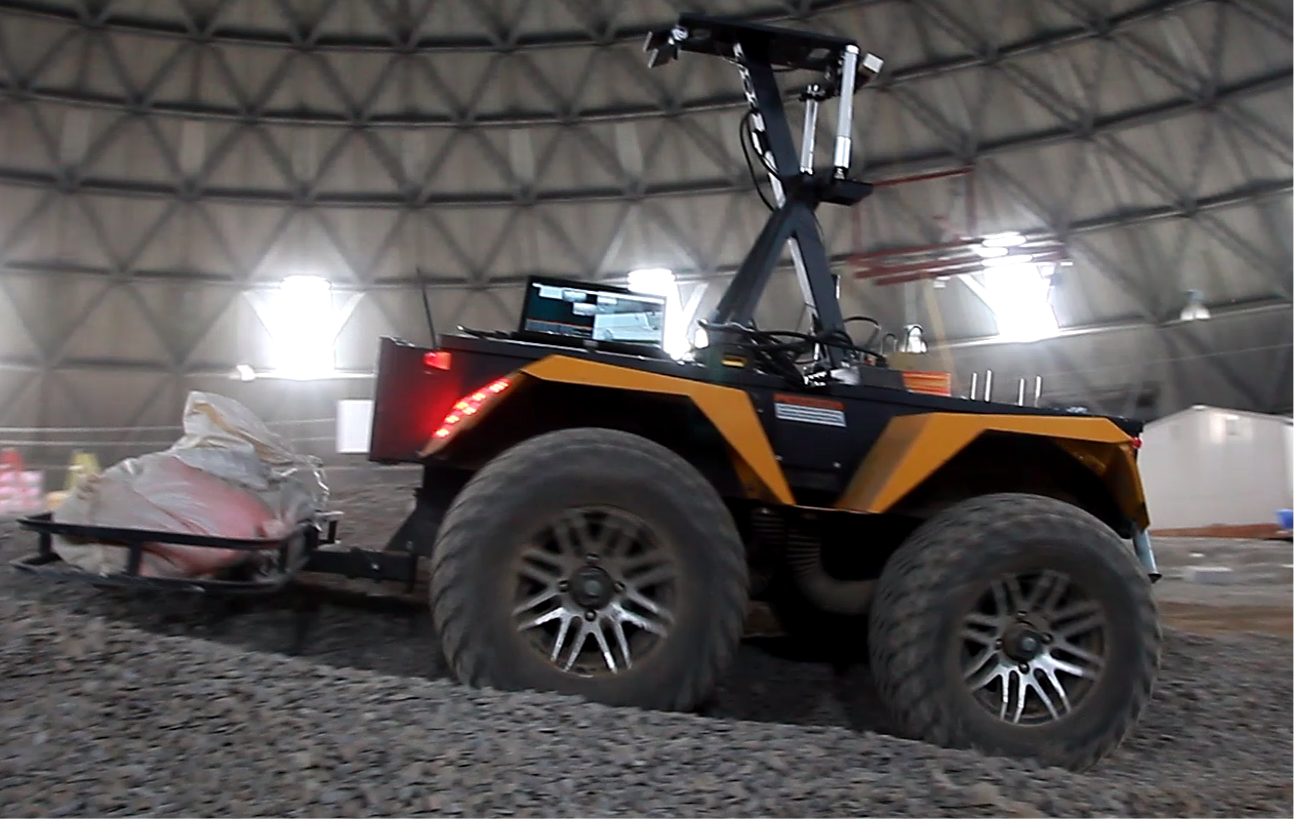}
	\vspace{-0.5cm}
	\caption{Clearpath Grizzly  in the \emph{loaded} configuration traversing a gravel mound at a target speed of 2.0 m/s with the proposed algorithm.}
	\label{fig:LoadedGrizzly}
\end{figure}

This section outlines how to apply our method to the unicycle ground robot pictured in Fig. \ref{fig:LoadedGrizzly}.

\subsection{Robot Model}
Let $\sysState = [x, y, \theta]^T$, the 2D position and heading of the robot,  $\actState = [v, \omega]^T$, the speed and turn rate of the robot, and  $\ctrl = [v^{cmd}, \omega^{cmd}]^T$, the commanded speed and turn rate of the robot. We assume that the dynamics of $\sysState$ are well approximated by a unicycle
\begin{align}
\label{eq:FirstOrderUnicycleDynamics}
\underbrace{\bbm x_{k+1}\\y_{k+1}\\ \theta_{k+1} \ebm}_{\sysState_{k+1}} = 
		\underbrace{\bbm x_{k}\\y_{k}\\ \theta_{k} \ebm}_{\sysState_{k}} + dt
											\underbrace{\bbm v_k\cos\theta_k \\ 
                    											  v_k \sin\theta_k \\
                                             \omega_k\ebm}_{\mb{f}(\cdot)},
\end{align}
which is of the form \eqref{eq:GeneralSystemDynamics}. For wBLR, we will model the dynamics of $\actState$ as
\begin{equation}
\label{eq:GrizzlyDynamcsBLR}
  \underbrace{\bbm
        v_{k+1} \\
        \omega_{k+1}
   \ebm}_{\actState_{k+1}}
    = 
    \underbrace{\bbm
         v_k \\ 
         \omega_k 
      \ebm}_{\mb{g}^0(\cdot)}
       + dt \underbrace{\bbm [v^{cmd}_k, v_k]\mb{w}_k^v + \eta_k^{v} \\ [\omega^{cmd}_k, \omega_k]\mb{w}_k^{\omega} + \eta_k^{\omega} \ebm}_{\mb{g}_k(\cdot)},
\end{equation}
which is of the form \eqref{eq:GeneralActuatorDynamics}.

\subsection{Ancillary Control Design for the Unicycle with First Order Actuator Dynamics}
\label{sec:AncillaryCtrlForUnicycle}
The ancillary controller is meant to reduce uncertainty growth over the prediction horizon. For the unicycle, lateral uncertainty growth (which is constrained) depends on heading uncertainty and speed. Keeping uncertainty in these states low therefore keeps the lateral uncertainty low reducing the amount that the constraints are tightened (see \eqref{eq:TightenedStateConstraints}). With a linear feedback controller on the heading and speed error, the speed and turn rate dynamics become:
\begin{align}
\label{eq:FirstOrderUnicycleDynamicsWithKthAndKvAncillaryControllers}
\bbm v_{k+1}\\ \omega_{k+1} \ebm &=
   \bbm v_{k}\\ \omega_{k} \ebm + dt \bbm
                                             [v^{cmd}_{k} + K_{v} e^v_k, v_k]^T\mb{w}^v\\
                                             [\omega^{cmd}_{k} + K_{\theta}e^{\theta}_{k}, \omega_k]^T\mb{w}^{\omega} \ebm
\end{align}
where $e^{(\cdot)}_k={(\cdot)}_{k}-\bar{{(\cdot)}}_{k}$ is the difference between the state ${(\cdot)_k}$ and the predicted mean at time step $k$. These controllers keep the system close to the predicted speed and heading. 

\section{EXPERIMENTS}

Experiments were conducted on a 900\, kg Clearpath Grizzly skid-steer ground robot shown in Fig. \ref{fig:LoadedGrizzly}. First, we compare the predictive performance of a GP to our proposed method on a dataset with varied payload and terrain type. Second, we demonstrate the effectiveness of each component of our algorithm in closed loop. Finally, we demonstrate the path tracking performance of our algorithm at high speed on a 175\,m off-road course.

\subsection{Implementation}

Our algorithm was implemented in C++ on an Intel i7 2.70 GHz 8 core processor with 16 GB of RAM. Our controller relies on a vision-based system, Visual Teach and Repeat \cite{paton2017expanding}, for localization, which runs on the same laptop. The controller runs at 10\,Hz with a three second look-ahead discretized by 30 points. The optimization problem \eqref{eq:CostFcnCost}-\eqref{eq:CostFcnInputConstraints} is solved as a sequential quadratic program and re-linearized three times, taking an average of 70\,ms to compute the control. The model updates (Sec. \ref{sec:ShortTermLearning} and \ref{sec:LongTermLearning}) are executed at every time step. 

We consider the last three seconds of data (30 samples) from the live run for $\mathcal{D}_n^-$.  The penalties on lag, contouring, heading, speed, and turn rate error are 50, 200, 200, 2, and 2 respectively. The penalties on commanded speed, turn rate, and reference speed from their references are 1, 1, and 50 respectively. The penalties on rate of change of commands in the same order are 10, 15, and 5. The maximum lateral error is 2 m, $r^c$ is 1, and the ancillary controller gains are both $-5$. The prior strength, $n_0$, was set to 100. For the high speed experiment, we increased the penalty on commanded turning acceleration from 15 to 20 to achieve smoother performance on the rough terrain.

\subsection{Model Predictive Performance Comparison}

In order to evaluate the suitability of the proposed method for predictive control, we evaluate the predictive performance of the proposed method (Sec. \ref{sec:ShortTermLearning} and \ref{sec:LongTermLearning}) to a context-aware GP (c.f. \cite{McKinnon2018ExpRec}, except we learn the actuator dynamics and not an additive model error) with fixed hyperparameters (GP-Fixed-Rec) and with hyperparmeters optimized using MLE and a sliding window of the last 100 datapoints (GP-MLE-Rec). We consider the rotational dynamics because they differ the most between configurations.

We compare the model predictions given the inputs that were actually applied to the vehicle over the MPC prediction horizon to the actual state of the vehicle recorded at the corresponding times. To measure the accuracy of the prediction of the mean, we use the Multi-Step RMS Error (M-RMSE) over this horizon. To measure the accuracy of the model uncertainty estimate, we use the Multi-Step RMS Z-score (M-RMSZ) over this horizon:
\begin{equation}
    M-RMSZ_k = \sqrt{\frac{1}{H} \sum_{q=0}^{H-1}\left(\frac{\omega_{k+q+1} - \bar{\omega}(\mb{x}_{k+q})}{\sigma^{\omega}(\mb{x}_{k+q})}\right)^2},
\end{equation}
where $\bar{\omega}(\mb{x}_k)$ is the predicted mean value of $\omega_k$ given the predicted $\mb{x}_k$ and $H$ is the number of timesteps in the prediction horizon. To generate the predictions, we use the controls inputs that were actually applied to the vehicle. An accurate model uncertainty estimate is important to ensure that the probability of violating the chance constraints formulated in Sec. \ref{sec:ConstraintTightening}  is kept at an acceptable level, specified by $\boldsymbol\epsilon^{\mb{z}}$ and $\boldsymbol\epsilon^{\mb{u}}$. We consider an M-RMSZ between -0.5 and 1.5 to be acceptable. If this value exceeds 2.0, the model uncertainty estimate is overconfident which could lead to violation of the chance constraints.


\begin{figure}
\centering
\scriptsize
\def\svgwidth{0.475\textwidth}
\graphicspath{{figs/Experiments/}}
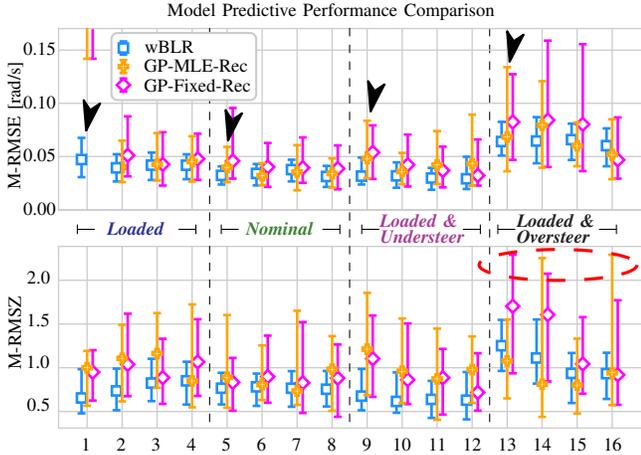
\vspace{-0.25cm}
\caption{A comparison of M-RMSE and M-RMSZ  for the rotational dynamics with the vehicle in four different configurations for two context-aware GP-based methods and the proposed method. The error bars indicate the 25th and 75th percentiles and the marker indicates the median. The 65\,m path traversed sand, gravel, and concrete. Runs 1-4  are in the {\color{Blue}\emph{\textbf{Loaded}}} configuration, with 6 gravel bags in the rear of the Grizzly (see Fig. \ref{fig:LoadedGrizzly}), runs 6-8 are with the vehicle in the {\color{OliveGreen}\emph{\textbf{Nominal}}} configuration (no modification), runs 9-12 are with the vehicle in the  {\color{Mulberry}\emph{\textbf{Loaded \& Understeer}}} configuration, where it is loaded and the turn rate commands are multiplied by 0.7, runs 13-16 are in the {\color{Black}\emph{\textbf{Loaded \& Oversteer}}} configuration, where the vehicle is loaded and the turn rate commands are multiplied by 1.2. The black arrows indicate the first time the vehicle is driven in a new operating condition. The red circle indicates where the GP-based methods were over-confident, frequently producing M-RMSZ values above 2.0.}
	\label{fig:RegressionPerformanceLongTerm}
\end{figure}

Figure \ref{fig:RegressionPerformanceLongTerm} compares the proposed method to GP-Fixed-Rec and GP-MLE-Rec. The proposed method consistently achieves lower M-RMSE, especially during run~1 before the GP-based methods have data, and the first time the system encounters a new configuration as indicated by the black arrows. This is the proposed method is able to incorporate relevant data from the current run using \emph{fast adaptation}. While online hyperparameter optimization generally improves the M-RMSE, it causes the GP overfit in the most challenging scenario, {\color{Black}\emph{\textbf{Loaded \& Oversteer}}}, which can be inferred by the M-RMSZ value exceeding 2.0 during runs 14 and 16. In contrast, the proposed method is much more consistent and the M-RMSZ stays between 0.5 and 1.5, indicating the model has a reasonable estimate of model uncertainty.

\subsection{Closed Loop Tracking Performance Comparison}
\label{sec:ClosedLoopComparison}

\begin{figure}
\centering
\scriptsize
\def\svgwidth{0.475\textwidth}
\graphicspath{{figs/Experiments/}}
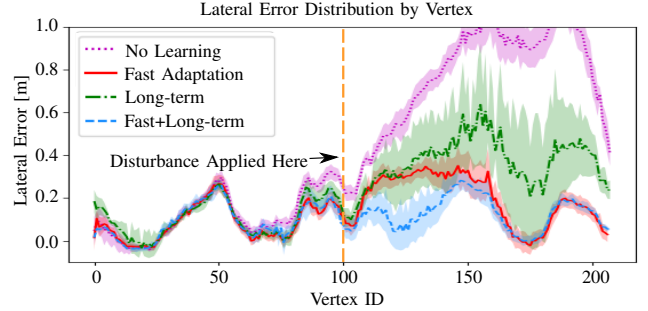
\vspace{-0.25cm}
\caption{This figure shows the closed-loop performance of the controller when we introduce a large, repetitive disturbance at vertex 100 by multiplying the turn rate commands by 0.5 after this point. This introduces a large, repeatable disturbance such as one might expect if the vehicle was traversing a patch of ice. The solid line indicates the median lateral tracking error over eight runs and the shaded region indicates the 50th and 75th percentiles. The proposed method with both long-term and fast adaptation learning achieves the lowest error and fastest convergence. No learning is when the controller uses a fixed wBLR model to compute the controls.}
	\label{fig:LatErrorVids}
\end{figure}

To demonstrate the impact of each component of our method in closed-loop and show that it can adapt to repetitive model errors, we drive the vehicle around two laps of a circular course and apply an artificial disturbance by multiplying the turn rate commands by 0.5 at the start of the second lap (vertex 100 in Fig. \ref{fig:LatErrorVids}). Physically, this may be similar to the vehicle getting a flat tyre or losing power in one motor. We compare the tracking performance of each component of our algorithm over eight repeats of the path. For this experiment, the desired speed was 2\,m/s.

\begin{figure}
\centering
\scriptsize
\def\svgwidth{0.475\textwidth}
\graphicspath{{figs/Experiments/}}
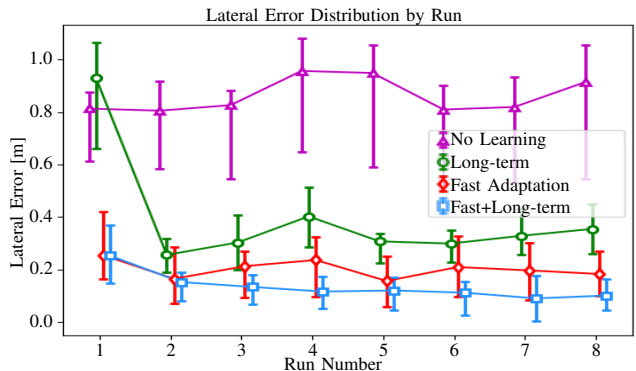
\vspace{-0.25cm}
\caption{Figure showing the 25th, 50th, and 75th percentiles of lateral error after the large, repetitive disturbance described in Sec. \ref{sec:ClosedLoopComparison} was applied. This figure shows that the proposed algorithm was able to quickly adapt to the disturbance and that the combination of fast adaptation (Sec. \ref{sec:ShortTermLearning}) and long-term learning (Sec. \ref{sec:LongTermLearning}) achieves the best performance. No learning is when the controller uses a fixed, prior model to compute the controls. The horizontal position of each point is offset slightly for clarity.}
	\label{fig:LatErrorRuns}
\end{figure}

Figure \ref{fig:LatErrorVids} shows that all methods achieve similar performance before the disturbance is applied because the model for all methods was a good representation of the vehicle dynamics over this portion of the path. After this point, the non-learning controller incurs a large lateral error because the model is no longer accurate. Long-term learning (Sec. \ref{sec:LongTermLearning}) similarly incurs a large path tracking error on the first run (see Figure \ref{fig:LatErrorRuns}) since there are no previous runs with experience. However, after the first run, it improves greatly  but then converges slowly because it is constantly working against a static prior (the same model used for the non-learning comparison), that is incorrect after the disturbance is applied. When fast adaptation (Sec. \ref{sec:ShortTermLearning}) is enabled, the controller incurs a large tracking error at the moment the disturbance is applied but adapts quickly to the new robot dynamics to achieve low error as expected. When both fast adaptation and long-term learning are enabled, the fast adaptation keeps the prior close to the true dynamics such that the long-term learning is able to reduce the transient error by leveraging data from the upcoming section of the path. This combination achieves the lowest path tracking error and the fastest convergence (see Fig. \ref{fig:LatErrorRuns}).

\subsection{High Speed Tracking Performance}

Finally, we evaluated the performance of our controller on a 175\,m off-road course with tight turns and fast straights. The desired speed was 3\,m/s and the controller achieved an average speed of 1.6\,m/s with a top speed of 2.7\,m/s and a RMS lateral error of 0.25\,m. This is a 60\% improvement over our previous work, where the controller achieved an average speed around 1.0\,m/s on pavement \cite{McKinnon2018ExpRec}.

\begin{figure}
\centering
\scriptsize
\def\svgwidth{0.475\textwidth}
\graphicspath{{figs/Experiments/}}
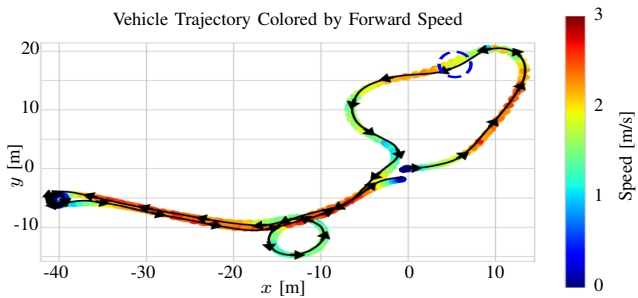
\vspace{-0.25cm}
\caption{This figure shows the path taken by the vehicle on five traverses of a 175\,m course. The  direction of travel is indicated by the black arrows. The maximum path tracking error is 0.7\,m when the controller cuts a corner (dashed blue circle). The vehicle was in the $\emph{Nominal}$ configuration.}
	\label{fig:ControlCostComparison}
\end{figure}

%% file: figs/Experiments/BLR_GPR-opthparams-recdata_GPD-fixedhparams-optdata.pdf_tex
\begingroup%
  \makeatletter%
  \providecommand\color[2][]{%
    \errmessage{(Inkscape) Color is used for the text in Inkscape, but the package 'color.sty' is not loaded}%
    \renewcommand\color[2][]{}%
  }%
  \providecommand\transparent[1]{%
    \errmessage{(Inkscape) Transparency is used (non-zero) for the text in Inkscape, but the package 'transparent.sty' is not loaded}%
    \renewcommand\transparent[1]{}%
  }%
  \providecommand\rotatebox[2]{#2}%
  \ifx\svgwidth\undefined%
    \setlength{\unitlength}{484.55039063bp}%
    \ifx\svgscale\undefined%
      \relax%
    \else%
      \setlength{\unitlength}{\unitlength * \real{\svgscale}}%
    \fi%
  \else%
    \setlength{\unitlength}{\svgwidth}%
  \fi%
  \global\let\svgwidth\undefined%
  \global\let\svgscale\undefined%
  \makeatother%
  \begin{picture}(1,0.70242559)%
    \put(0,0){\includegraphics[width=\unitlength]{BLR_GPR-opthparams-recdata_GPD-fixedhparams-optdata.pdf}}%
    \put(0.58372195,0.35329058){\color[rgb]{0,0,0}\makebox(0,0)[lb]{\smash{{\color{Mulberry}\emph{\textbf{Loaded \&}}}}}}%
    \put(0.5839138,0.32719478){\color[rgb]{0,0,0}\makebox(0,0)[lb]{\smash{{\color{Mulberry}\emph{\textbf{Understeer}}}}}}%
    \put(0.01871823,0.12862504){\color[rgb]{0,0,0}\rotatebox{90}{\makebox(0,0)[lb]{\smash{M-RMSZ}}}}%
    \put(0.02238176,0.37171456){\color[rgb]{0,0,0}\makebox(0,0)[lb]{\smash{0.00}}}%
    \put(0.02250591,0.45549584){\color[rgb]{0,0,0}\makebox(0,0)[lb]{\smash{0.05}}}%
    \put(0.02238176,0.53927713){\color[rgb]{0,0,0}\makebox(0,0)[lb]{\smash{0.10}}}%
    \put(0.02250591,0.62305836){\color[rgb]{0,0,0}\makebox(0,0)[lb]{\smash{0.15}}}%
    \put(0.01756139,0.41211941){\color[rgb]{0,0,0}\rotatebox{90}{\makebox(0,0)[lb]{\smash{M-RMSE [rad/s]}}}}%
    \put(0.11363561,0.00016368){\color[rgb]{0,0,0}\makebox(0,0)[lb]{\smash{1}}}%
    \put(0.16908049,0.00016368){\color[rgb]{0,0,0}\makebox(0,0)[lb]{\smash{2}}}%
    \put(0.22403441,0.0003273){\color[rgb]{0,0,0}\makebox(0,0)[lb]{\smash{3}}}%
    \put(0.27960908,0.00016368){\color[rgb]{0,0,0}\makebox(0,0)[lb]{\smash{4}}}%
    \put(0.33461943,0.00047965){\color[rgb]{0,0,0}\makebox(0,0)[lb]{\smash{5}}}%
    \put(0.38999095,0.0003273){\color[rgb]{0,0,0}\makebox(0,0)[lb]{\smash{6}}}%
    \put(0.44522703,0.00031603){\color[rgb]{0,0,0}\makebox(0,0)[lb]{\smash{7}}}%
    \put(0.50069444,0.0003273){\color[rgb]{0,0,0}\makebox(0,0)[lb]{\smash{8}}}%
    \put(0.55606602,0.0003273){\color[rgb]{0,0,0}\makebox(0,0)[lb]{\smash{9}}}%
    \put(0.60333967,0.0003273){\color[rgb]{0,0,0}\makebox(0,0)[lb]{\smash{10}}}%
    \put(0.65954065,0.00016368){\color[rgb]{0,0,0}\makebox(0,0)[lb]{\smash{11}}}%
    \put(0.71435919,0.00016368){\color[rgb]{0,0,0}\makebox(0,0)[lb]{\smash{12}}}%
    \put(0.76940907,0.0003273){\color[rgb]{0,0,0}\makebox(0,0)[lb]{\smash{13}}}%
    \put(0.82445888,0.00016368){\color[rgb]{0,0,0}\makebox(0,0)[lb]{\smash{14}}}%
    \put(0.88009567,0.0003273){\color[rgb]{0,0,0}\makebox(0,0)[lb]{\smash{15}}}%
    \put(0.93525841,0.0003273){\color[rgb]{0,0,0}\makebox(0,0)[lb]{\smash{16}}}%
    \put(0.24912739,0.68486421){\color[rgb]{0,0,0}\makebox(0,0)[lb]{\smash{Model Predictive Performance Comparison}}}%
    \put(0.15441158,0.34024268){\color[rgb]{0,0,0}\makebox(0,0)[lb]{\smash{{\color{Blue}\emph{\textbf{Loaded}}}}}}%
    \put(0.3731129,0.34024268){\color[rgb]{0,0,0}\makebox(0,0)[lb]{\smash{{\color{OliveGreen}\emph{\textbf{Nominal}}}}}}%
    \put(0.21324621,0.62678904){\color[rgb]{0,0,0}\makebox(0,0)[lb]{\smash{wBLR}}}%
    \put(0.21232074,0.59668007){\color[rgb]{0,0,0}\makebox(0,0)[lb]{\smash{GP-MLE-Rec}}}%
    \put(0.21232074,0.56635662){\color[rgb]{0,0,0}\makebox(0,0)[lb]{\smash{GP-Fixed-Rec}}}%
    \put(0.02799978,0.06268526){\color[rgb]{0,0,0}\makebox(0,0)[lb]{\smash{0.5}}}%
    \put(0.02722668,0.12952717){\color[rgb]{0,0,0}\makebox(0,0)[lb]{\smash{1.0}}}%
    \put(0.02735082,0.19636908){\color[rgb]{0,0,0}\makebox(0,0)[lb]{\smash{1.5}}}%
    \put(0.02785306,0.26321099){\color[rgb]{0,0,0}\makebox(0,0)[lb]{\smash{2.0}}}%
    \put(0.79932716,0.35318901){\color[rgb]{0,0,0}\makebox(0,0)[lb]{\smash{{\color{Black}\emph{\textbf{Loaded \&}}}}}}%
    \put(0.79930462,0.32729635){\color[rgb]{0,0,0}\makebox(0,0)[lb]{\smash{{\color{Black}\emph{\textbf{Oversteer}}}}}}%
  \end{picture}%
\endgroup%

%% file: figs/Experiments/LateralErrorDistributionCircles_by_vertex.pdf_tex
\begingroup%
  \makeatletter%
  \providecommand\color[2][]{%
    \errmessage{(Inkscape) Color is used for the text in Inkscape, but the package 'color.sty' is not loaded}%
    \renewcommand\color[2][]{}%
  }%
  \providecommand\transparent[1]{%
    \errmessage{(Inkscape) Transparency is used (non-zero) for the text in Inkscape, but the package 'transparent.sty' is not loaded}%
    \renewcommand\transparent[1]{}%
  }%
  \providecommand\rotatebox[2]{#2}%
  \ifx\svgwidth\undefined%
    \setlength{\unitlength}{444.85795898bp}%
    \ifx\svgscale\undefined%
      \relax%
    \else%
      \setlength{\unitlength}{\unitlength * \real{\svgscale}}%
    \fi%
  \else%
    \setlength{\unitlength}{\svgwidth}%
  \fi%
  \global\let\svgwidth\undefined%
  \global\let\svgscale\undefined%
  \makeatother%
  \begin{picture}(1,0.53989363)%
    \put(0,0){\includegraphics[width=\unitlength]{LateralErrorDistributionCircles_by_vertex.pdf}}%
    \put(0.30452128,0.46187504){\color[rgb]{0,0,0}\makebox(0,0)[lb]{\smash{Lateral Error Distribution by Vertex}}}%
    \put(0.03061102,0.16184458){\color[rgb]{0,0,0}\rotatebox{90}{\makebox(0,0)[lb]{\smash{Lateral Error [m]}}}}%
    \put(0.04412334,0.43660907){\color[rgb]{0,0,0}\makebox(0,0)[lb]{\smash{1.0}}}%
    \put(0.04473976,0.3685091){\color[rgb]{0,0,0}\makebox(0,0)[lb]{\smash{0.8}}}%
    \put(0.04469235,0.3004091){\color[rgb]{0,0,0}\makebox(0,0)[lb]{\smash{0.6}}}%
    \put(0.04455536,0.23230905){\color[rgb]{0,0,0}\makebox(0,0)[lb]{\smash{0.4}}}%
    \put(0.04507168,0.16780571){\color[rgb]{0,0,0}\makebox(0,0)[lb]{\smash{0.2}}}%
    \put(0.04472923,0.09610907){\color[rgb]{0,0,0}\makebox(0,0)[lb]{\smash{0.0}}}%
    \put(0.31941222,0.04110572){\color[rgb]{0,0,0}\makebox(0,0)[lb]{\smash{50}}}%
    \put(0.50840117,0.04110572){\color[rgb]{0,0,0}\makebox(0,0)[lb]{\smash{100}}}%
    \put(0.70466071,0.04110572){\color[rgb]{0,0,0}\makebox(0,0)[lb]{\smash{150}}}%
    \put(0.13382499,0.04110572){\color[rgb]{0,0,0}\makebox(0,0)[lb]{\smash{0}}}%
    \put(0.90510165,0.04110572){\color[rgb]{0,0,0}\makebox(0,0)[lb]{\smash{200}}}%
    \put(0.18645103,0.39575242){\color[rgb]{0,0,0}\makebox(0,0)[lb]{\smash{No Learning}}}%
    \put(0.18493369,0.36016426){\color[rgb]{0,0,0}\makebox(0,0)[lb]{\smash{Fast Adaptation}}}%
    \put(0.18617707,0.32394387){\color[rgb]{0,0,0}\makebox(0,0)[lb]{\smash{Long-term}}}%
    \put(0.18493369,0.2855037){\color[rgb]{0,0,0}\makebox(0,0)[lb]{\smash{Fast+Long-term}}}%
    \put(0.47735693,0.00630666){\color[rgb]{0,0,0}\makebox(0,0)[lb]{\smash{Vertex ID}}}%
    \put(0.16228934,0.22289113){\color[rgb]{0,0,0}\makebox(0,0)[lb]{\smash{Disturbance Applied Here}}}%
  \end{picture}%
\endgroup%

%% file: figs/Experiments/LateralErrorDistributionCircles_by_run.pdf_tex
\begingroup%
  \makeatletter%
  \providecommand\color[2][]{%
    \errmessage{(Inkscape) Color is used for the text in Inkscape, but the package 'color.sty' is not loaded}%
    \renewcommand\color[2][]{}%
  }%
  \providecommand\transparent[1]{%
    \errmessage{(Inkscape) Transparency is used (non-zero) for the text in Inkscape, but the package 'transparent.sty' is not loaded}%
    \renewcommand\transparent[1]{}%
  }%
  \providecommand\rotatebox[2]{#2}%
  \ifx\svgwidth\undefined%
    \setlength{\unitlength}{491.75561523bp}%
    \ifx\svgscale\undefined%
      \relax%
    \else%
      \setlength{\unitlength}{\unitlength * \real{\svgscale}}%
    \fi%
  \else%
    \setlength{\unitlength}{\svgwidth}%
  \fi%
  \global\let\svgwidth\undefined%
  \global\let\svgscale\undefined%
  \makeatother%
  \begin{picture}(1,0.57549551)%
    \put(0,0){\includegraphics[width=\unitlength]{LateralErrorDistributionCircles_by_run.pdf}}%
    \put(0.31636958,0.55077541){\color[rgb]{0,0,0}\makebox(0,0)[lb]{\smash{Lateral Error Distribution by Run}}}%
    \put(0.0275788,0.16089704){\color[rgb]{0,0,0}\rotatebox{90}{\makebox(0,0)[lb]{\smash{Lateral Error [m]}}}}%
    \put(0.1423515,0.02262165){\color[rgb]{0,0,0}\makebox(0,0)[lb]{\smash{1}}}%
    \put(0.25407664,0.02262165){\color[rgb]{0,0,0}\makebox(0,0)[lb]{\smash{2}}}%
    \put(0.36529084,0.02285206){\color[rgb]{0,0,0}\makebox(0,0)[lb]{\smash{3}}}%
    \put(0.47715106,0.02262165){\color[rgb]{0,0,0}\makebox(0,0)[lb]{\smash{4}}}%
    \put(0.58842396,0.02306649){\color[rgb]{0,0,0}\makebox(0,0)[lb]{\smash{5}}}%
    \put(0.70007275,0.02285206){\color[rgb]{0,0,0}\makebox(0,0)[lb]{\smash{6}}}%
    \put(0.81158063,0.02283617){\color[rgb]{0,0,0}\makebox(0,0)[lb]{\smash{7}}}%
    \put(0.92332931,0.0225644){\color[rgb]{0,0,0}\makebox(0,0)[lb]{\smash{8}}}%
    \put(0.70112553,0.35491382){\color[rgb]{0,0,0}\makebox(0,0)[lb]{\smash{No Learning}}}%
    \put(0.70089512,0.28279816){\color[rgb]{0,0,0}\makebox(0,0)[lb]{\smash{Fast Adaptation}}}%
    \put(0.70089512,0.31927699){\color[rgb]{0,0,0}\makebox(0,0)[lb]{\smash{Long-term}}}%
    \put(0.70089512,0.24778092){\color[rgb]{0,0,0}\makebox(0,0)[lb]{\smash{Fast+Long-term}}}%
    \put(0.0385053,0.0682859){\color[rgb]{0,0,0}\makebox(0,0)[lb]{\smash{0.0}}}%
    \put(0.03888705,0.15106771){\color[rgb]{0,0,0}\makebox(0,0)[lb]{\smash{0.2}}}%
    \put(0.03831148,0.23384956){\color[rgb]{0,0,0}\makebox(0,0)[lb]{\smash{0.4}}}%
    \put(0.03846418,0.31663139){\color[rgb]{0,0,0}\makebox(0,0)[lb]{\smash{0.6}}}%
    \put(0.03851704,0.39941319){\color[rgb]{0,0,0}\makebox(0,0)[lb]{\smash{0.8}}}%
    \put(0.03782989,0.48219503){\color[rgb]{0,0,0}\makebox(0,0)[lb]{\smash{1.0}}}%
    \put(0.43910926,0.00046073){\color[rgb]{0,0,0}\makebox(0,0)[lb]{\smash{Run Number}}}%
  \end{picture}%
\endgroup%

%% file: figs/Experiments/TopDownLongRun-short.pdf_tex
\begingroup%
  \makeatletter%
  \providecommand\color[2][]{%
    \errmessage{(Inkscape) Color is used for the text in Inkscape, but the package 'color.sty' is not loaded}%
    \renewcommand\color[2][]{}%
  }%
  \providecommand\transparent[1]{%
    \errmessage{(Inkscape) Transparency is used (non-zero) for the text in Inkscape, but the package 'transparent.sty' is not loaded}%
    \renewcommand\transparent[1]{}%
  }%
  \providecommand\rotatebox[2]{#2}%
  \ifx\svgwidth\undefined%
    \setlength{\unitlength}{583.5581543bp}%
    \ifx\svgscale\undefined%
      \relax%
    \else%
      \setlength{\unitlength}{\unitlength * \real{\svgscale}}%
    \fi%
  \else%
    \setlength{\unitlength}{\svgwidth}%
  \fi%
  \global\let\svgwidth\undefined%
  \global\let\svgscale\undefined%
  \makeatother%
  \begin{picture}(1,0.44931907)%
    \put(0,0){\includegraphics[width=\unitlength]{TopDownLongRun-short.pdf}}%
    \put(0.16934765,0.42797603){\color[rgb]{0,0,0}\makebox(0,0)[lb]{\smash{Vehicle Trajectory Colored by Forward Speed}}}%
    \put(0.98594501,0.14990216){\color[rgb]{0,0,0}\rotatebox{89.73742914}{\makebox(0,0)[lb]{\smash{Speed [m/s]}}}}%
    \put(0.40190117,0.00573166){\color[rgb]{0,0,0}\makebox(0,0)[lb]{\smash{$x$ [m]}}}%
    \put(0.06045482,0.02867779){\color[rgb]{0,0,0}\makebox(0,0)[lb]{\smash{-40}}}%
    \put(0.1975692,0.02867779){\color[rgb]{0,0,0}\makebox(0,0)[lb]{\smash{-30}}}%
    \put(0.33468359,0.02867779){\color[rgb]{0,0,0}\makebox(0,0)[lb]{\smash{-20}}}%
    \put(0.62902134,0.02867779){\color[rgb]{0,0,0}\makebox(0,0)[lb]{\smash{0}}}%
    \put(0.75674487,0.02867779){\color[rgb]{0,0,0}\makebox(0,0)[lb]{\smash{10}}}%
    \put(0.47453978,0.02867779){\color[rgb]{0,0,0}\makebox(0,0)[lb]{\smash{-10}}}%
    \put(0.02222762,0.17042932){\color[rgb]{0,0,0}\rotatebox{89.4912893}{\makebox(0,0)[lb]{\smash{$y$ [m]}}}}%
    \put(0.01364741,0.10373616){\color[rgb]{0,0,0}\makebox(0,0)[lb]{\smash{-10}}}%
    \put(0.0310146,0.19623842){\color[rgb]{0,0,0}\makebox(0,0)[lb]{\smash{0}}}%
    \put(0.0243656,0.28874068){\color[rgb]{0,0,0}\makebox(0,0)[lb]{\smash{10}}}%
    \put(0.02488571,0.38124292){\color[rgb]{0,0,0}\makebox(0,0)[lb]{\smash{20}}}%
    \put(0.92698894,0.00767625){\color[rgb]{0,0,0}\makebox(0,0)[lb]{\smash{0}}}%
    \put(0.92717636,0.15000647){\color[rgb]{0,0,0}\makebox(0,0)[lb]{\smash{1}}}%
    \put(0.92727476,0.29247256){\color[rgb]{0,0,0}\makebox(0,0)[lb]{\smash{2}}}%
    \put(0.92696551,0.43507456){\color[rgb]{0,0,0}\makebox(0,0)[lb]{\smash{3}}}%
  \end{picture}%
\endgroup%